\documentclass[letterpaper, 10 pt, conference]{ieeeconf}
\IEEEoverridecommandlockouts
\usepackage{cite}
\usepackage{amsmath,amssymb,amsfonts}
\usepackage{algorithm}
\usepackage{algpseudocode}
\usepackage{graphicx}
\usepackage{comment}
\usepackage{url}
\usepackage{hyperref}
\usepackage{textcomp}
\usepackage{xcolor}
\usepackage[nolist]{acronym}
\usepackage{colortbl}
\usepackage{siunitx}
\usepackage{booktabs}

\usepackage{tikz, pgfplots, pgfplotstable}
\usetikzlibrary{positioning, arrows.meta, fit, backgrounds, calc, shapes.geometric, decorations.markings, patterns, decorations.pathreplacing, fadings, shadows, decorations.pathreplacing, decorations.markings}
\usepgfplotslibrary{statistics}
\pgfplotsset{compat=1.18}
\usetikzlibrary{external}
\usepackage{eso-pic}
\usepackage{makecell}   
\usepackage{threeparttable} 
\usepackage{pifont}     

\usepackage[acronym, nomain]{glossaries}
\newacronym{vlm}{VLM}{ Vision Language Model}
\newacronym{vla}{VLA}{ Vision Language Action}
\newacronym{sota}{SOTA}{State-of-the-Art}
\newacronym{mllm}{MLLM}{Multimodal Large Language Model}
\newacronym{ad}{AD}{Autonomous Driving}
\newacronym{odd}{ODD}{Operational Design Domain}
\newacronym{e2e}{E2E}{End-to-End}
\newacronym{llm}{LLM}{Large Language Model}
\newacronym{cot}{CoT}{Chain of Thought}
\newacronym{pinn}{PINN}{Physics-Informed Neural Network}
\newacronym{pikc}{PIKC}{Physics-Informed Kinematic Consistency}
\newacronym{homoscedastic}{Homoscedastic}{Homoscedastic Uncertainty Weighting}
\newacronym{heteroscedastic}{Heteroscedastic}{Heteroscedastic Uncertainty Weighting}
\newacronym{crossentropy}{Cross-Entropy}{Cross-Entropy Loss}
\newacronym{mse}{MSE}{Mean Squared Error}
\newacronym{kl}{KL}{Kullback-Leibler Divergence}
\newacronym{bce}{BCE}{Binary Cross-Entropy}
\newacronym{ce}{CE}{Cross-Entropy}
\newacronym{bev}{BEV}{Bird's Eye View}
\newacronym{fpv}{FPV}{First Person View}
\newacronym{rms}{RMS}{Root Mean Square}
\newacronym{vqa}{VQA}{Visual Question Answering}
\newacronym{mlp}{MLP}{Multi-Layer Perceptron}
\newacronym{ade}{ADE}{Average Displacement Error}
\newacronym{fde}{FDE}{Final Displacement Error}
\newacronym{psr}{PSR}{Planning Success Rate}
\newacronym{kce}{KCE}{Kinematic Consistency Error}
\newacronym{mr}{MR}{Miss Rate}
\newacronym{cr}{CR}{CommonRoad}

\usetikzlibrary{shapes.geometric, arrows, positioning}

\definecolor{Blue}{RGB}{0,101,189}
\definecolor{White}{RGB}{255,255,255}
\definecolor{Black}{RGB}{0,0,0}
\definecolor{Gray}{RGB}{156,157,159}
\definecolor{GrayLight}{RGB}{217,218,219}
\definecolor{Green}{RGB}{162,173,0}
\definecolor{Orange}{HTML}{E37222}
\definecolor{Yellow}{RGB}{252, 231, 46}
\definecolor{Red}{RGB}{196,7,27}

\tikzstyle{block} = [rectangle, rounded corners, minimum width=3cm, minimum height=1cm,text centered, draw=black, fill=blue!20]
\tikzstyle{input} = [ellipse, minimum width=2cm, minimum height=1cm, text centered, draw=black, fill=green!20]
\tikzstyle{output} = [ellipse, minimum width=2cm, minimum height=1cm, text centered, draw=black, fill=red!20]
\tikzstyle{arrow} = [thick,->,>=stealth]

\def\BibTeX{{\rm B\kern-.05em{\sc i\kern-.025em b}\kern-.08em
    T\kern-.1667em\lower.7ex\hbox{E}\kern-.125emX}}
\begin{document}

\title{StyleVLA: Driving Style-Aware Vision Language Action Model for\\ Autonomous Driving}

\author{Yuan Gao$^{1}$, Dengyuan Hua$^{1}$, Mattia Piccinini$^{1}$, Finn Rasmus Schäfer$^{1}$, Korbinian Moller$^{1}$, Lin Li$^{2}$, Johannes Betz$^{1}$ \\ \url{https://anonymous-paper-2026.github.io/StyleVLA/}
\thanks{$^{1}$ Y. Gao, D. Hua, M. Piccinini, F. Schäfer, K. Moller, J. Betz are with the Professorship of Autonomous Vehicle Systems, TUM School of Engineering and Design, Technical University of Munich, 85748 Garching, Germany; Munich Institute of Robotics and Machine Intelligence (MIRMI)}
\thanks{$^{2}$ L. Li is with the School of Mechanical
and Aerospace Engineering, Nanyang Technological University
}
}

\maketitle
\AddToShipoutPictureBG*{%
    \AtPageLowerLeft{%
        \hspace*{\dimexpr(\paperwidth-\textwidth)/2}%
        \raisebox{0.8cm}{%
            \fbox{\parbox{\dimexpr\textwidth-2\fboxsep-2\fboxrule}{%
                \centering\scriptsize
                This work has been submitted to the IEEE for possible publication. Copyright may be transferred without notice, after which this version may no longer be accessible.%
            }}%
        }%
    }%
}
\begin{abstract}
\glspl{vlm} are transforming intelligent systems by bridging the gap between visual perception and linguistic reasoning. In \gls{ad}, this synergy has catalyzed the development of \gls{vla} models, which aim to translate high-level multimodal understanding into actionable driving behaviors, typically represented as future trajectories. However, current \gls{vla} models predominantly focus on generating generic collision-free trajectories. While collision avoidance is a fundamental requirement, adapting to diverse driving styles (e.g., sporty, comfortable) is essential for personalized user experiences. Furthermore, they often treat trajectory generation as a naive token prediction task, leading to kinematically infeasible actions. To address these limitations, we present \textbf{StyleVLA}, a physics-informed \gls{vla} framework that generates diverse, physically plausible driving behaviors. We introduce a novel hybrid loss function that integrates a physics-informed kinematic consistency constraint with a continuous regression head, enhancing the physical feasibility of generated trajectories. To train the StyleVLA model based on Qwen3-VL 4B, we construct a large-scale instruction dataset containing over 1.2k scenarios with 76k \gls{bev} and 42k \gls{fpv} samples, featuring ground-truth trajectories for five distinct driving styles and natural-language instructions. Extensive experiments demonstrate that our 4B-parameter StyleVLA model significantly outperforms proprietary models (e.g., Gemini-3-Pro) and \gls{sota} \gls{vla} models. Using a composite driving score that measures success rate, physical feasibility, and adherence to user-specified driving styles, StyleVLA achieves 0.55 on \gls{bev} and 0.51 on \gls{fpv}, compared to 0.32 and 0.35 for Gemini-3-Pro. This finding highlights that a specialized, physics-informed, lightweight model can surpass closed-source models on domain-specific tasks.
\end{abstract}

\section{Introduction}
Foundation models, particularly \glspl{llm} and \acrfull{vlm}, have revolutionized artificial intelligence by demonstrating remarkable reasoning and generalization capabilities across diverse domains, from natural language processing to \acrfull{ad}~\cite{gao2025survey}. By leveraging massive-scale pre-training on internet-scale data, these models can understand complex multi-modal contexts and perform tasks with minimal zero-shot adaptation~\cite{gao2026foundation}. In the realm of \gls{ad}, this paradigm shift has catalyzed the development of \acrfull{vla} models. These models aim to transcend the limitations of traditional rule-based systems and classical \gls{e2e} architectures by using the multi modal reasoning of \glspl{vlm} for decision-making~\cite{huang2025drivegpt}. 

\begin{figure}[tp]
    \centering
    \includegraphics[width=0.85\linewidth]{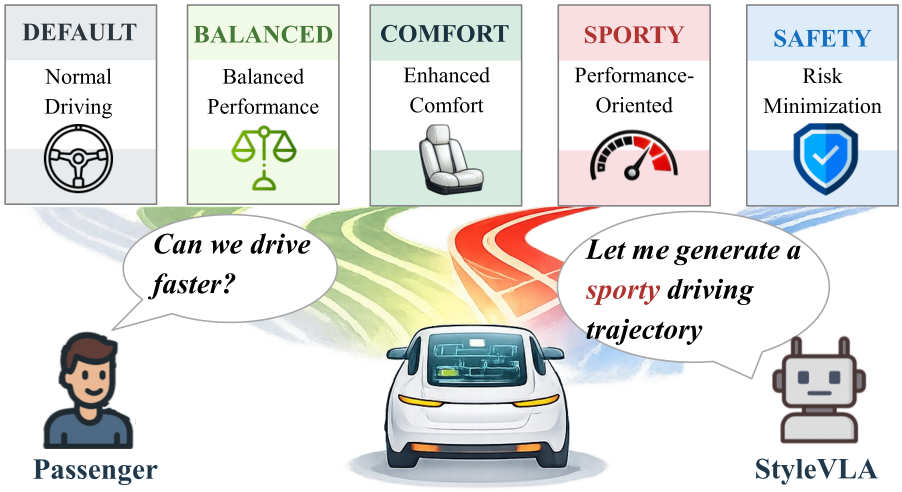}
    \caption{Concept of StyleVLA: Enabling driving style-aware trajectory generation via \gls{vla} model. Our framework yields diverse driving styles (Default, Balanced, Comfort, Sporty, Safety) in response to user instructions.}
    \label{fig:concept}
\end{figure}

Specifically, existing \gls{vla} models prioritize collision avoidance, neglecting the heterogeneity of human driving preferences. As illustrated in \autoref{fig:concept}, real-world driving requires adapting to diverse driving styles, such as sporty or comfort-oriented behaviors, based on user intent. This deficiency stems from the lack of large-scale datasets with ground-truth trajectories for diverse driving styles. To address this, we propose \textbf{StyleVLA}, a driving style-aware \gls{vla} model for trajectory generation in autonomous driving. This model leverages physics-informed supervision to generate diverse, physically plausible driving behaviors. To enable this, we also construct the accompanying \textbf{StyleVLA} dataset, which provides ground-truth supervision for distinct driving styles (Default, Balanced, Comfort, Sporty, Safety).

\subsection{Related Work}
\subsubsection{Classical E2E Autonomous Driving}
\label{sec:e2e_planning}
Recent literature increasingly favors \gls{e2e} \gls{ad} to address the limitations of modular pipelines, aiming to jointly optimize the entire driving stack by directly mapping raw sensor inputs to control outputs via neural networks~\cite{chen2024end}. While early behavioral cloning approaches~\cite{bojarski2020nvidia} suffered from covariate shift, classical unified architectures like UniAD~\cite{hu2023planning} and VAD~\cite{jiang2023vad} have advanced the field through hierarchical query-based Transformers and efficient vectorized representations. However, for these traditional architectures~\cite{chen2024end}, handling rare, long-tail scenarios and generalizing to diverse driving behaviors remain significant challenges. This has motivated the evolution towards \gls{vla} models, which integrate \glspl{vlm} into the \gls{e2e} framework to provide enhanced reasoning capabilities for more effectively addressing these complexities.

\subsubsection{VLMs in Autonomous Driving}
\label{sec:vlm_planning}
\glspl{vlm} integrate visual encoders with language model backbones, enabling robust cross-modal reasoning capabilities~\cite{gao2026foundation}. Leveraging their pre-trained nature and adaptability, \glspl{vlm} have been applied to path generation via three paradigms: \gls{e2e} Planners (often referred to as \gls{vla} models), Hybrid Systems, and Teacher-Student Distillation~\cite{oksuz2025foun}. 
While \gls{vla} models map sensor inputs directly to trajectory outputs by formulating planning as a multimodal generation task (e.g., EMMA~\cite{hwang2025emma}) or using latent feature injection (e.g., VLP~\cite{pan2024vlp}), challenges such as quantization noise persist. Recent frameworks have addressed these limitations through distinct mechanisms: OpenDriveVLA~\cite{zhou2025opendrivevla} and Orion~\cite{fu2025orion} improve feature alignment and long-term context integration, while Alpamayo~\cite{wang2025alpamayo} and SimLingo~\cite{renz2025simlingo} focus on enhancing reasoning-action consistency and instruction adherence. To address latency, Hybrid Systems (e.g., DriveVLM~\cite{tian2025drivevlm}) decouple reasoning from control, using a slower \gls{vlm} for high-level decision making and a faster traditional motion planner for trajectory generation. Alternatively, Distillation approaches (e.g., VLM-AD~\cite{xu2024vlm}) train smaller student models to predict \gls{vlm}-derived insights offline. Despite these advances, the integration of diverse driving styles into \gls{vla} frameworks remains underexplored. Most existing models assume a single driving policy, limiting their ability to adapt trajectory generation to heterogeneous user preferences.

\subsubsection{Autonomous Driving Datasets}
\label{sec:datasets}
The development of style-aware \gls{vla} models is fundamentally constrained by the quality and diversity of available training data. High-quality datasets are essential for developing robust \gls{ad} systems. A recent foundation model survey~\cite{gao2026foundation} highlights several high-impact datasets, including Waymo Open~\cite{sun2020waymoopen}, nuScenes~\cite{caesar2020nuscenes}, HighD~\cite{Krajewski2018highd}, and DRAMA~\cite{malla2023drama}. These datasets offer rich multimodal sensor data, including \acrfull{bev} and \acrfull{fpv} images, combined with detailed 2D/3D annotations and ego-trajectory information. However, they lack explicit annotations and diverse data distributions that represent heterogeneous driving styles. This limitation restricts the ability of current \glspl{vla} models to learn and execute personalized driving strategies.

\subsubsection{Style-Aware Autonomous Driving}
While safety and efficiency are paramount for \gls{ad}, adapting to diverse driving styles is crucial for user acceptance and comfort. Recent works have explored personalized driving behaviors to address this need. MAVERIC~\cite{schrum2024maveric} learns user-specific driving-style embeddings from demonstrations and predicts parameters for low-level controllers. StyleDrive~\cite{hao2025styledrive} introduces a benchmark for evaluating driving-style awareness in \gls{e2e} driving using coarse driving style labels (aggressive, normal, and conservative). However, these approaches primarily focus on controller-level personalization or predefined style categories, rather than enabling flexible trajectory generation conditioned on user preferences.

\subsection{Critical Summary}
\label{sec:critical_summary}
To the best of our knowledge, existing literature is limited by the following aspects:
\begin{itemize}
    \item \textbf{Limited driving-style diversity in existing datasets.} 
    Current \gls{ad} datasets~\cite{sun2020waymoopen, caesar2020nuscenes, Krajewski2018highd, malla2023drama} provide rich multimodal perception data but lack explicit annotations and distributions that capture diverse driving styles (e.g., cautious or sporty), hindering research on personalized \gls{ad}.
    \item \textbf{Lack of style-controllable trajectory generation.} 
    Existing \gls{vla} models are typically trained on homogeneous driving data and therefore lack mechanisms to condition trajectory generation on user-specified driving styles, resulting in generic driving behaviors.
    \item \textbf{Lack of physics-informed trajectory supervision.}
    Many \gls{vla} models treat trajectory generation as a token prediction task~\cite{mao2023gpt, sima2024drivelm} or use external decoders~\cite{wang2025alpamayo, fu2025orion}, often without explicitly modeling vehicle kinematic constraints.
\end{itemize}

\subsection{Contributions}
\label{sec:contributions}

The key contributions of this paper are as follows:
\begin{itemize}
    \item  We present the StyleVLA dataset ($1,216$ scenarios, $76,030$ \gls{bev} samples and $42,084$ \gls{fpv} samples), featuring trajectories with five distinct driving styles (Default, Balanced, Comfort, Sporty, Safety) and language instructions. This dataset enables training and evaluation of style-aware \gls{vla} models for personalized \gls{ad}.
     \item  We propose a physics-informed \gls{vla} model fine-tuning framework for generating driving style-aware trajectories. It integrates standard \gls{ce} loss with an auxiliary \gls{mlp} regression head and a physics-informed kinematic loss to improve trajectory feasibility and style adherence when fine-tuning a 4B \gls{vlm}, outperforming zero-shot \glspl{vlm} and \gls{sota} \gls{vla} models on unseen data.
    \item  We conduct a large-scale evaluation of off-the-shelf \glspl{vlm} and \gls{sota} \gls{vla} methods on the StyleVLA dataset across \gls{bev} and \gls{fpv} domains, revealing their limitations in style-aware trajectory generation.
\end{itemize}

\section{Methodology}
\label{ch:methodology}
\begin{figure*}[!t]
    \centering
    \includegraphics[width=\textwidth]{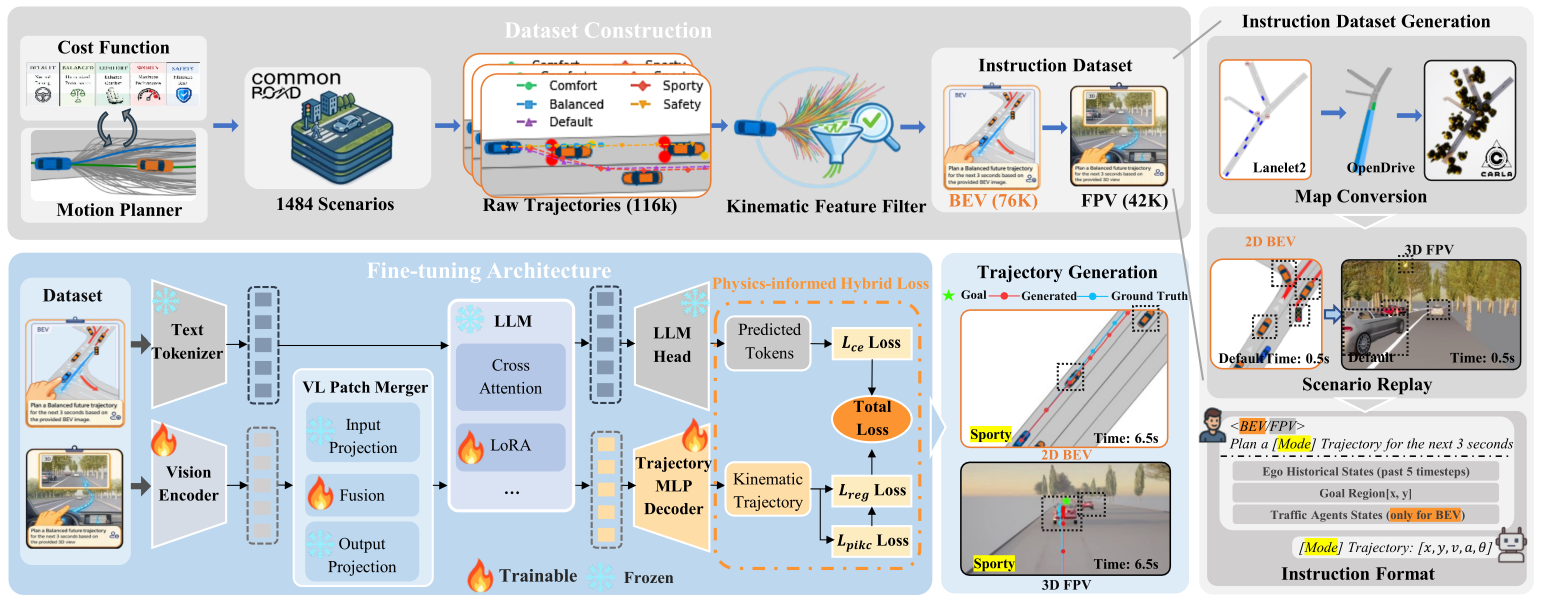} 
    \caption{Overview of the StyleVLA framework. \textbf{Top (Dataset Construction):} A motion planner generates style-specific ground-truth trajectories to create multimodal instruction samples. \textbf{Instruction Dataset Generation:} Details the instruction generation process and 3D scenario replay in CARLA. \textbf{Bottom (Fine-tuning Architecture):}  The model predicts trajectory tokens using only an LLM head conditioned on visual context and language prompts. During training, an auxiliary MLP decoder maps the predicted tokens to continuous kinematic trajectories for physics-informed supervision. Training uses a physics-informed hybrid loss ($\mathcal{L}_\mathrm{total}$) combining cross-entropy ($\mathcal{L}_\mathrm{ce}$), regression ($\mathcal{L}_\mathrm{reg}$), and kinematic consistency ($\mathcal{L}_\mathrm{pikc}$). \textbf{Trajectory Generation:} Shows the model's application in both 2D BEV and 3D FPV domains. }
    \label{fig:framework_overview}
\end{figure*}

This section details the methodology for developing StyleVLA (\autoref{fig:framework_overview}). First, we describe the construction of the StyleVLA dataset, where we generate ground-truth trajectories for distinct driving styles using a multi-objective motion planner. Second, we explain the creation of multimodal instruction datasets for both \gls{bev} and \gls{fpv} domains, pairing visual contexts with style-specific language instructions. Finally, we present our fine-tuning framework, which employs a physics-informed hybrid loss to train the \gls{vla} model for precise and kinematically consistent trajectory generation. 

\subsection{StyleVLA Dataset Construction}
\label{sec:StyleVLA_dataset_construction}
To generate trajectories with diverse, custom driving styles, we employ the open-source sampling-based motion planner Frenetix~\cite{trauth2024frenetix} within the \gls{cr} framework~\cite{althoff2017commonroad}. Different driving styles are realized by adapting the motion planner's multi-objective cost function to prioritize style-specific metrics such as comfort or safety.
\begin{table}[b]
    \centering
    \caption{Weights across Driving Styles, for the cost function \eqref{eq:cost_fun_planner} of the Frentix motion planner used for dataset generation.}
    \label{tab:cost_weights}
    \begin{threeparttable}
    \resizebox{\columnwidth}{!}{%
    \renewcommand{\arraystretch}{1.1}
    \setlength{\tabcolsep}{3pt}
    \begin{tabular}{lccccc}
        \toprule
        \textbf{Cost Term} & \textbf{Comfort} & \textbf{Balanced} & \textbf{Sporty} & \textbf{Safety} & \textbf{Default} \\
        \midrule
        \rowcolor{gray!20} \textit{Kinematic Constraints} ($\mathbf{w}_{\mathrm{kin}}$) & & & & & \\
        Longitudinal Jerk ($w_\mathrm{j,lon}$) & \textbf{0.80} & 0.50 & 0.25 & 0.40 & 0.20 \\
        Lateral Jerk ($w_\mathrm{j,lat}$) & \textbf{0.80} & 0.50 & 0.25 & 0.40 & 0.20 \\
        Velocity Offset ($w_\mathrm{v}$) & 0.30 & 0.60 & \textbf{1.00} & 0.30 & 1.00 \\
        Distance to obstacles ($w_\mathrm{obs}$) & 0.30 & 0.80 & 0.60 & \textbf{2.00} & 0.00 \\
        \midrule
        \rowcolor{gray!20} \textit{External Perception} ($\mathbf{w}_{\mathrm{ext}}$) & & & & & \\
        Phantom Risk ($w_\mathrm{pm}$)~\cite{Moller2025_Occlusion} & 3.0 & 5.0 & 4.0 & \textbf{8.0} & 5.0 \\
        Visibility Seeking ($w_\mathrm{ve}$)~\cite{Trauth2023_Occlusion} & 0.0 & 0.5 & 0.8 & \textbf{1.5} & 0.0 \\
        \bottomrule
    \end{tabular}
    }
\par
    \end{threeparttable}
\end{table}

\textbf{Stage 1: Cost Function Design.} At each planning step, the motion planner generates a set of candidate trajectories $\mathcal{X}$ by sampling target end states in the curvilinear coordinate frame $(s,d)$ along a given reference path. The sampling spans $m$ lateral displacements $d$ and $n$ longitudinal velocities $\dot{s}$, yielding up to $m \times n$ trajectory candidates $\xi$ per step. 

Each trajectory $\xi$ is represented as a sequence of kinematic state vectors:
\begin{equation}
\xi = \{ \mathbf{s}_t \}_{t=1}^{T}, \quad \mathbf{s}_t = [x_t, y_t, v_t, a_t, \theta_t]^\top \in \mathbb{R}^5,
\label{eq:trajectory_state_vector}
\end{equation}
where $(x_t, y_t)$ denote the ego-vehicle position, $v_t$ the velocity, $a_t$ the longitudinal acceleration, and $\theta_t$ the heading angle at timestep $t$. $T$ denotes the maneuver duration.

Each candidate trajectory $\xi$ is subsequently subjected to a kinematic feasibility check enforcing bounds on acceleration, curvature, and yaw rate, with infeasible samples being discarded. The remaining feasible trajectories $\mathcal{X}_\mathrm{feas}$ are then evaluated using a driving style-specific cost function $J_k$ for each style $k \in \mathcal{D}$, where $\mathcal{D} = \{\textit{Comfort}, \textit{Sporty}, \textit{Safe}, \textit{Balanced}, \textit{Default}\}$ denotes the set of considered driving styles. The cost for a trajectory $\xi$ under driving style $k$ is defined as
\begin{equation}
J_k(\xi) = \mathbf{w}_{\mathrm{kin}, k}^\top \mathbf{C}_{\mathrm{kin}}(\xi) + \mathbf{w}_{\mathrm{ext}, k}^\top \mathbf{C}_{\mathrm{ext}}(\xi),
\label{eq:cost_fun_planner}
\end{equation}
which balances internal kinematic costs $\mathbf{C}_{\mathrm{kin}}$ (e.g., jerk or deviation from the desired velocity), weighted by $\mathbf{w}_{\mathrm{kin},k}$, and external perceptual costs $\mathbf{C}_{\mathrm{ext}}$ (e.g., occlusion risk), weighted by $\mathbf{w}_{\mathrm{ext},k}$. \autoref{tab:cost_weights} lists the corresponding weight configurations for each driving style $k \in \mathcal{D}$. Weights are designed according to the behavioral role of each cost term and tuned to produce distinct driving styles.

Among the feasible trajectories $\mathcal{X}_\mathrm{feas}$, the trajectory with the lowest cost $J_k$ is selected as the style-conditioned output $\xi_k^*$. By adjusting the weight vectors $\mathbf{w}_{\mathrm{kin},k}$ and $\mathbf{w}_{\mathrm{ext},k}$ for each style, the same sampling pool produces qualitatively different driving behaviors. Specifically, \textit{Comfort Mode} prioritizes passenger comfort by penalizing jerk, \textit{Sporty Mode} favors faster progress by penalizing deviations from the desired velocity, \textit{Safety Mode} enforces larger spatial buffers around obstacles, and \textit{Balanced Mode} represents a moderate trade-off between these objectives. The \textit{Default Mode} serves as a baseline configuration. 

With the five driving styles $k\in \mathcal{D}$ defined, we next prepare the scenario corpus on which Frenetix~\cite{trauth2024frenetix} is executed. We use $1,484$ scenarios drawn from the \gls{cr} scenario database~\cite{althoff2017commonroad}. The \gls{cr} database comprises a large collection of traffic scenarios with road networks represented as Lanelet2 maps~\cite{poggenhans2018lanelet2}, enabling an accurate representation of real-world road structures. The scenarios originate from $14$ countries (e.g., Germany: $597$, Greece: $258$, Poland: $245$) and include diverse traffic scenarios (e.g., urban intersections, roundabouts, and highways) under varying traffic conditions.

The corpus comprises $53,457$ dynamic agents and \SI{11.2}{\hour} of driving data, with an average scenario duration of \SI{27.13}{\second}. Running all five styles across each scenario at a replanning frequency of \SI{0.5}{\second} yields $116,400$ planning instances. Each instance represents a single time step, consisting of a \gls{bev} image and the corresponding ground-truth trajectory generated by the motion planner. However, environmental constraints can override style preferences, resulting in identical behaviors across styles (e.g., \textit{Sporty} vs. \textit{Comfort} in dense traffic). To ensure clear supervision, we filter out ambiguous samples in which kinematics do not clearly reflect the assigned style.
\begin{table}[t]
    \centering
    \caption{Mean Kinematic Features and Distribution of Filtered Trajectories by Style}
    \label{tab:style_stats_mean}
    \begin{threeparttable}
    \resizebox{\columnwidth}{!}{%
    \renewcommand{\arraystretch}{1.1}
    \setlength{\tabcolsep}{3pt}
        \begin{tabular}{lc cccc}
            \toprule
            \textbf{Style Label} & \textbf{Samples} & \textbf{Avg Velocity} & \textbf{RMS Accel} & \textbf{RMS Jerk} & \textbf{Path Length} \\
             & (Count / \%) & ($m/s$) & ($m/s^2$) & ($m/s^3$) & ($m$) \\
            \midrule
            Balanced & 14,102 (18.5\%) & 7.15 & 0.588 & 0.750 & 24.44 \\
            Comfort & 13,766 (18.1\%) & 7.21 & 0.585 & \textbf{0.727} & 24.53 \\
            Default & 17,684 (23.3\%) & 6.80 & 0.486 & 0.794 & 23.31 \\
            Sporty & 14,790 (19.5\%) & \textbf{7.32} & 0.558 & 0.780 & \textbf{25.13} \\
            Safety & 15,688 (20.6\%) & 6.39 & 0.583 & 0.746 & 21.44 \\
            \bottomrule
        \end{tabular}
    }
    \end{threeparttable}
\end{table}

\textbf{Stage 2: Dataset Filtering.}
For every trajectory $\xi_k$ with assigned style $k\in \mathcal{D}$, we compute a fixed, scalar summary by aggregating kinematic statistics over all timesteps into a constant feature vector $\mathbf{f}_{\xi,k}$:
\begin{equation}
\mathbf{f}_{\xi,k} = [ \bar{v},\ \sigma_v,\ a_{\mathrm{rms}},\ |a|_{\mathrm{max}},\ j_{\mathrm{rms}},\ \sigma_j ]^\top,
\end{equation}
where $\bar{v}$ is the mean velocity, $\sigma_v$ the standard deviation of velocity, $a_{\mathrm{rms}}$ the \gls{rms} acceleration, $|a|_{\mathrm{max}}$ the peak absolute acceleration, $j_{\mathrm{rms}}$ the \gls{rms} jerk, and $\sigma_j$ the standard deviation of jerk, all computed as scalar aggregates over the full trajectory duration. $\mathbf{f}_{\xi,k}$ is thus a single constant vector per trajectory.

To define the ground truth distribution for each style $k \in \mathcal{D}$, we collect $\{\mathbf{f}_{\xi,k}\}$ over all raw trajectories labeled $k$ and fit a multivariate Gaussian $\mathcal{N}(\mu_k, \Sigma_k)$, where $\mu_k \in \mathbb{R}^6$ is the style-specific mean vector and $\Sigma_k \in \mathbb{R}^{6\times6}$ is the covariance matrix. We estimate $(\mu_k, \Sigma_k)$ using the Minimum Covariance Determinant (MCD) estimator~\cite{Hubert_2017}, which fits the Gaussian to the densest subset of the data and is therefore robust to outliers in the raw pool. We then measure how well $\mathbf{f}_{\xi,k}$ conforms to the fitted distribution $\mathcal{N}(\mu_k, \Sigma_k)$ using the Mahalanobis distance $D_M$:
{\small
\begin{equation}
D_M(\xi_k) = \sqrt{(\mathbf{f}_{\xi,k} - \mu_k)^T \Sigma_k^{-1} (\mathbf{f}_{\xi,k} - \mu_k)}.
\end{equation}
}
This distance is mapped to a probabilistic conformance score $S \in [0, 100]$ via the Chi-squared CDF with $d=6$ degrees of freedom:
{\small
\begin{equation}
S(\xi_k) = 100 \cdot \left( 1 - \chi^2_{\text{cdf}}(D_M^2(\xi_k),\ d) \right).
\end{equation}
}
We retain samples with a conformance score $S(\xi_k) > 80$, corresponding to the lowest \SI{20}{\percent} of the Chi-squared distribution, a common statistical threshold for selecting samples that closely conform to the reference distribution. The filtering process yields a refined dataset of $76,030$ planning instances across $1,216$ scenarios. \autoref{tab:style_stats_mean} presents the mean kinematic characteristics of the final filtered BEV dataset, illustrating the quantitative distinctions preserved between styles.

%
%
\subsection{Instruction Dataset Generation (BEV Domain)}
\label{sec:bev_instruction}

To transform raw kinematic data into a format suitable for fine-tuning \glspl{vlm}, we constructed a multimodal instruction dataset (\autoref{fig:framework_overview}). This process involves pairing visual context with linguistic instructions and historical state data, creating a rich supervised learning target.

To ensure broad compatibility and leverage a proven instruction-following structure, we adopt the widely used LLaVA \gls{vqa} conversation format~\cite{liu2023visual}. Each \gls{vqa} sample is structured into three key components aligned with the LLaVA format: the \texttt{image} field (Visual Input), the \texttt{human} message (Human Instruction), and the \texttt{gpt} message (Model Response).


\noindent \textbf{1) Visual Input:} The primary spatial context is provided by a single \gls{bev} image generated by the \gls{cr} environment. It captures a \SI{30}{\meter} radius local map where dynamic obstacles are distinguished by geometric shapes and colors, and road topology is defined by boundaries and markings.
    
\noindent \textbf{2) Human Instruction:} The user query integrates multi-modal context into a structured text prompt. It includes: (1) \textit{Ego Vehicle History}, a \SI{0.5}{\second} sequence of ego-states sampled at \SI{10}{\hertz}; (2) \textit{Traffic Agents States}, listing the kinematic states of the $10$ nearest neighbor agents within the \SI{30}{\meter} radius; (3) \textit{Goal Region}, defining the target position; and (4) \textit{Style Command}, a natural language instruction (e.g., "Plan a trajectory with [Mode] driving style...") that directs the model to adopt specific behavioral characteristics. Given the \gls{vlm}'s sensitivity to prompts, we use DSPy\footnote{\url{https://dspy.ai/}} to automatically optimize this instruction.
    
\noindent \textbf{3) Model Response:} The model generates the future trajectory as a structured JSON object, covering a horizon of \SI{3}{\second} (standard) or \SI{5}{\second} (extended) at \SI{2}{\hertz}. The response encodes the full kinematic state vector $\mathbf{s}_t = [x_t, y_t, v_t, a_t, \theta_t]$ \eqref{eq:trajectory_state_vector} for each timestamp, rather than just positions, to support physics-informed loss calculation.

\subsection{Instruction Dataset Generation (FPV Domain)}
\label{sec:fpv_instruction}
To extend StyleVLA to 3D, we use the CARLA simulator and extend the \gls{bev} images with \gls{fpv} camera images for realistic \gls{e2e} driving as shown in \autoref{fig:framework_overview}.

\subsubsection{Map Conversion and Scenario Replay}
We convert \gls{cr} scenario maps to the OpenDRIVE format\footnote{\url{https://www.asam.net/standards/detail/opendrive/}}
preserving lane topology, junctions, traffic signs, and traffic lights. The resulting OpenDRIVE maps are directly compatible with CARLA.
To enhance visual realism, we implement a procedural landscape generation system that populates the environment with vegetation. Trees are spawned \SIrange{4}{18}{\meter} from road edges using waypoint-based terrain elevation calculations, ensuring proper ground alignment.

For each scenario, we perform a complete resimulation in CARLA with synchronized video recording. The camera is mounted on the vehicle roof in a \gls{fpv} configuration following the nuScenes front-facing camera convention, positioned forward and above the vehicle center with a \SI{5}{\degree} downward tilt to capture the road ahead while maintaining horizon visibility. We replay the driving-style ground-truth trajectories by spawning both the ego vehicle and all traffic participants. At each simulation timestep, we update vehicle positions and orientations using coordinate transformation utilities that convert \gls{cr} states to CARLA instances. To match \gls{cr} obstacles to appropriate CARLA vehicle models, we employ a dimensional-similarity-matching system. The system maintains a database of CARLA vehicle dimensions and selects blueprints based on length, width, and height similarity, ensuring accurate visual representation of different vehicle types (e.g., cars, trucks, and buses). 

\subsubsection{Quality Control and Instruction Generation}
To ensure data integrity, we implemented a two-stage filtering pipeline. First, an automated validation step discards frames with rendering failures (e.g., black screens), or map conversion errors (e.g., off-road spawning due to coordinate misalignment). Second, a human-in-the-loop verification removes scenarios with ambiguous visual cues or incomplete traffic participant spawning. This process yielded a final dataset of \textbf{42,084 high-quality instances} from successfully replayed scenarios. 

We modify the \textbf{Human Instruction} to enforce implicit perception and vision-based driving. Unlike the \gls{bev} setting, where traffic-agent states are provided in the prompt, the \gls{fpv} setting is vision-only: we omit external traffic states to prevent shortcut learning (i.e., relying on provided states instead of perception). The prompt contains only: (1) \textit{Ego-Vehicle History}, (2) \textit{Goal Point}, and (3) \textit{Style Command}.

%
\subsection{Fine-Tuning StyleVLA}
\label{sec:finetuning}
To enhance the capability of \gls{vla} models to generate diverse driving-style trajectories, we fine-tune a \gls{vlm} on our \textbf{StyleVLA instruction dataset}. We adopt Qwen3-VL-4B~\cite{bai2025qwen3} as our base model due to its strong multimodal reasoning capabilities and efficient parameter count, making it practical for deployment on edge platforms~\cite{gao2026foundation}. To make fine-tuning feasible on consumer-grade hardware, we employ QLoRA (low-rank adaptation with 4-bit quantization), freezing language model weights and training lightweight adapter matrices in the attention and feed-forward layers (\autoref{fig:framework_overview}).

To bridge the gap between discrete semantic reasoning and continuous control, we introduce a physics-informed hybrid loss function. This objective jointly optimizes geometric accuracy and kinematic plausibility, improving trajectory feasibility compared to standard token-based prediction (\autoref{tab:ablation_combined}).
\subsubsection{Hybrid Loss Function Design}
\begin{figure}
    \centering
    \resizebox{0.9\linewidth}{!}{\input{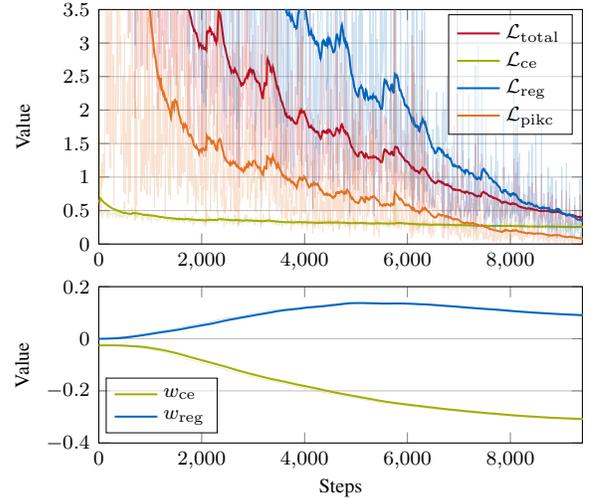}}
    \vspace{-4mm}
    \caption{Example training dynamics of StyleVLA fine-tuning on the \gls{fpv} instruction dataset. Top: loss terms ($\mathcal{L}_\mathrm{total}$, $\mathcal{L}_\mathrm{ce}$, $\mathcal{L}_\mathrm{reg}$, $\mathcal{L}_\mathrm{pikc}$). Bottom: learned log-variance parameters ($w_\mathrm{ce}$, $w_\mathrm{reg}$) that yield adaptive precision weights via $\exp(-w)$.}
    \label{fig:paramter}
\end{figure}

Standard \glspl{vlm} are trained using a \gls{ce} loss $\mathcal{L}_\mathrm{ce}$~\cite{liu2023visual}, treating trajectory generation as a next-token prediction task:
{\small
\begin{equation}
    \mathcal{L}_\mathrm{ce} = -\frac{1}{N} \sum_{i=1}^{N} \log P_\theta(\tau_{\mathrm{gt},i} \mid X_i, I_i),
    \end{equation}
}
where $N$ is the number of training samples. $\tau_{\mathrm{gt}, i}$ denotes the response with the target style trajectory (e.g., \SI{3}{\second} $\mathbf{s}_t = [x_t, y_t, v_t, a_t, \theta_t]$), $X_i$ the instruction prompt, $I_i$ the visual input, and $P_\theta(\cdot)$ the conditional probability distribution over output tokens predicted by the model with parameters $\theta$.
However, token-level classification discretizes continuous states, potentially introducing quantization error.
To address this, we introduce an auxiliary \gls{mlp} regression head attached to the Transformer's final hidden states (\autoref{fig:framework_overview}). This head projects the pooled semantic embedding of the response into a continuous sequence of kinematic states $\hat{\xi}_\mathrm{reg}$, allowing us to minimize geometric error against the ground truth $\xi_\mathrm{gt}$:
{\small
\begin{equation}
\mathcal{L}_\mathrm{reg} = || \hat{\xi}_\mathrm{reg} - \xi_\mathrm{gt} ||_2^2.
\label{eq:l_reg}
\end{equation}
}
The \gls{mlp} regression head is used only during training to provide physics-informed supervision. During inference, trajectory generation is performed solely by the \gls{llm} decoding head, which outputs structured trajectory tokens.
To balance the discrete \gls{ce} loss ($\mathcal{L}_\mathrm{ce}$) and the continuous regression loss ($\mathcal{L}_\mathrm{reg}$), which differ significantly in scale and convergence dynamics, we adopt a Homoscedastic Uncertainty Weighting strategy~\cite{kendall2018multi}. The unified hybrid loss objective is derived from maximizing the Gaussian likelihood:

\vspace{-2mm}
{\small
\begin{equation}
\mathcal{L}_\mathrm{hybrid} = \left( e^{-w_\mathrm{ce}} \mathcal{L}_\mathrm{ce} + \frac{1}{2} w_\mathrm{ce} \right) + \frac{1}{2} \left( e^{-w_\mathrm{reg}} \mathcal{L}_\mathrm{reg} + w_\mathrm{reg} \right),
\label{eq:l_hybrid}
\end{equation}
}

\noindent where $w_\mathrm{ce}$ and $w_\mathrm{reg}$ are learnable log-variance parameters. Their corresponding precision weights, $\exp(-w_\mathrm{ce})$ and $\exp(-w_\mathrm{reg})$, adaptively rescale $\mathcal{L}_\mathrm{ce}$ and $\mathcal{L}_\mathrm{reg}$ during fine-tuning: smaller $w$ increases a term's contribution. \autoref{fig:paramter} visualizes the loss curves (top) and the learned parameters (bottom) during StyleVLA fine-tuning on the \gls{fpv} instruction dataset.
%
%
\subsubsection{\gls{pikc}}
To ensure physical plausibility, we enforce a \gls{pikc} Loss ($\mathcal{L}_\mathrm{pikc}$). By predicting the full state vector $\mathbf{s}_t = [x_t, y_t, v_t, a_t, \theta_t]$ at each timestep, we derive the physically expected next position $(\hat{x}_{t+1}, \hat{y}_{t+1})$ based on the current state and discrete kinematic equations with time step $\Delta t$:
\begin{small}
\begin{equation}
\begin{aligned}
\hat{x}_{t+1} &= x_t + v_t \cos\theta_t \Delta t + 0.5\, a_t \cos\theta_t (\Delta t)^2, \\
\hat{y}_{t+1} &= y_t + v_t \sin\theta_t \Delta t + 0.5 \, a_t \sin\theta_t (\Delta t)^2.
\end{aligned}
\label{eq:kine_pred}
\end{equation}
\end{small}
The consistency loss is then defined as the error between the model's directly predicted next position $(x_{t+1}, y_{t+1})$ and this physically extrapolated one:

\vspace{-4mm}
\begin{small}
\begin{equation}
\mathcal{L}_\mathrm{pikc} = \frac{1}{T-1} \sum_{t=0}^{T-1} \left( || x_{t+1} - \hat{x}_{t+1} ||^2 + || y_{t+1} - \hat{y}_{t+1} ||^2 \right).
\end{equation}
\end{small}

\noindent This loss term does not depend on the ground truth data but instead operates on the internal consistency of the prediction itself, allowing the \gls{mlp} head to learn a differentiable function of the vehicle's kinematics. The final regression objective $\mathcal{L}_\mathrm{reg,total}$ combines the direct loss $\mathcal{L}_\mathrm{reg}$ \eqref{eq:l_reg} with the kinematic penalty:
\begin{small}
\begin{equation}
\mathcal{L}_\mathrm{reg, total} = {\boldsymbol{w}}_\mathrm{reg}^\top\mathcal{L}_\mathrm{reg} + w_\mathrm{pikc}\mathcal{L}_\mathrm{pikc},
\label{eq:l_reg_total}
\end{equation}
\end{small}
where $\boldsymbol{w}_\mathrm{reg}$ and $w_\mathrm{pikc}$ are fixed loss weights (see \autoref{tab:loss_hparams}). We assign higher weights to the position and kinematic term, while velocity and heading serve as auxiliary guides. This physics-informed regularization is integrated into $\mathcal{L}_\mathrm{hybrid}$ \eqref{eq:l_hybrid} by replacing the $\mathcal{L}_\mathrm{reg}$ term with $\mathcal{L}_\mathrm{reg, total}$.
The resulting training objective is
\begin{small}
\begin{equation}
\mathcal{L}_{\mathrm{total}} =
\mathcal{L}_{\mathrm{hybrid}}
\Big|_{\mathcal{L}_{\mathrm{reg}} \rightarrow \mathcal{L}_{\mathrm{reg,total}}}.
\end{equation}
\end{small}
An ablation study on how the physics-informed hybrid loss affects the performance of the fine-tuned model is presented in \autoref{subsec:experiment_1}.
\section{Results \& Discussion}\label{sec:results}
\subsection{Experimental Setup}
All experiments are performed on a Dell Alienware R15 equipped with an Intel i7-13700KF CPU, an NVIDIA RTX 4090 GPU with 24GB VRAM, and 128 GB of RAM.

\subsubsection{Evaluated Vision Language Models}
Our evaluation includes leading proprietary and open-source \glspl{vlm}. For the proprietary models, we use the following models: 
Gemini 2.5 Pro, Gemini 2.5 Flash, Gemini 3 Pro, and GPT 5 Nano. Regarding open-source models, we leverage LMDeploy\footnote{\url{https://github.com/InternLM/lmdeploy}} to deploy Qwen3-VL-4B, Qwen2.5-VL-7B, InternVL3-9B.
\begin{table}[t]
    \centering
    \caption{Hyperparameters used in fine-tuning}
    \label{tab:loss_hparams}
    \resizebox{0.85\columnwidth}{!}{%
    \renewcommand{\arraystretch}{1.1}
    \begin{tabular}{l p{0.6\columnwidth}}
        \toprule
        \textbf{Group} & \textbf{Value} \\
        \midrule
        \rowcolor{gray!15} \multicolumn{2}{l}{{\textit{Regression Loss Weights}}} \\
        $(\boldsymbol{w}_\mathrm{reg}, w_\mathrm{pikc})$ & $(2.0,\,2.0,\,0.5,\,0.5,\,0.5,\,1.5)$ \\
        \rowcolor{gray!15} \multicolumn{2}{l}{\textit{Training Hyperparameters (\gls{fpv})}} \\
        Backbone & Qwen3-VL-4B-Instruct \\
        Quantization & QLoRA, 4-bit \\
        LoRA $(r,\alpha,\mathrm{dropout})$ & $(256,\,512,\,0.05)$ \\
        Optimizer/schedule & AdamW (8-bit), cosine, warmup 0.03 \\
        Learning rate & $1\times10^{-4}$ (base) \\
        Component LRs & vision $2\times10^{-6}$, merger $1\times10^{-5}$, reg-head $1\times10^{-5}$, $w$-params $5\times10^{-5}$ \\
        Precision & bf16 \\
        \bottomrule
    \end{tabular}
    }
\end{table}
%
%
\subsubsection{Evaluation metrics}
We evaluate generated trajectory quality using standard metrics: \gls{ade} and \gls{fde}, computed on the 2D position sequence $(x_t,y_t)$ as the mean and final Euclidean distance to the ground truth.  We also report the \gls{psr}, defined as the percentage of trajectories with \gls{ade} $< \SI{1.0}{\meter}$, and the \gls{mr} for failures with \gls{fde} $> \SI{2.0}{\meter}$. Additionally, we introduce the \gls{kce} to quantify physical violations, calculated as the discrepancy between the model's output position at $t+1$ and the predicted position computed from \eqref{eq:kine_pred}. Finally, we report the inference time for trajectory generation.

To resolve trade-offs between disparate metrics, we propose a unified grading formula $\mathcal{S}_\mathrm{final} \in [0, 1]$ that prioritizes safety and success over raw precision:
{\small
\begin{equation}
    \mathcal{S}_\mathrm{final} = 0.35 S_\mathrm{succ} + 0.30 S_\mathrm{reach} + 0.20 S_\mathrm{acc} + 0.15 S_\mathrm{kin}
\end{equation}
}
where the components are defined as:
{\small
\begin{equation}
\begin{aligned}
    S_\mathrm{succ} &= \text{PSR}, \quad S_\mathrm{reach} = 1 - \text{MR}, \\
    S_\mathrm{acc} &= 0.4 e^{-\frac{\text{ADE}}{1.5}} + 0.6 e^{-\frac{\text{FDE}}{3.0}}, \\
    S_\mathrm{kin} &= 0.3 S_\mathrm{vel} + 0.3 S_\mathrm{head} + 0.4 S_\mathrm{consist},
\end{aligned}
\end{equation}
}
with the kinematic sub-scores given by:
{\small
\begin{equation}
\begin{aligned}
S_\mathrm{vel} &= \max(0, 1 - \text{MAE}_{v}/3.0), \\ 
S_\mathrm{head} &= \max(0, 1 - \text{MAE}_{\theta}/0.2), \\ 
S_\mathrm{consist} &= \max(0, 1 - \text{KCE}/0.5).
\end{aligned}
\end{equation}
}

\begin{figure*}[t]
    \centering
    \includegraphics[width=\textwidth]{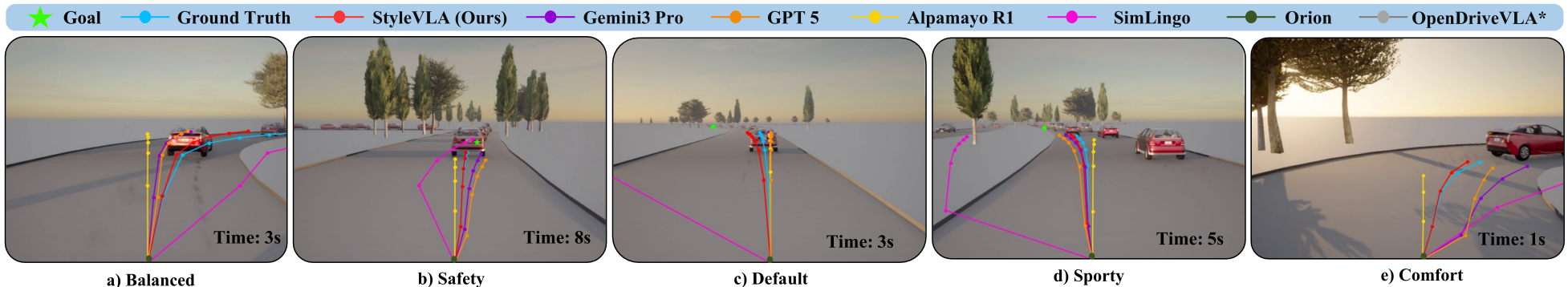}
     \vspace{-8mm}
    \caption{Qualitative comparison of style-conditioned trajectory generation under five driving styles (Default, Balanced, Comfort, Sporty, Safety). We visualize the goal, ground truth, and predicted trajectories from pretrained \glspl{vlm} and \gls{sota} baselines (see legend). "*" failed to generate trajectories.}
    \label{fig:carla_image}
\end{figure*}

\begin{table}[t]
    \centering
    \caption{Ablation on BEV StyleVLA (Qwen2.5-VL-7B, 3\,s horizon). Scaling data and adding physics-informed hybrid loss consistently improves generalization}
    \label{tab:ablation_combined}
    \resizebox{1\columnwidth}{!}{%
    \renewcommand{\arraystretch}{1.2}
    \begin{tabular}{l cccc}
        \toprule
        \textbf{Configuration} & \textbf{ADE} (m) $\downarrow$ & \textbf{FDE} (m) $\downarrow$ & \textbf{PSR} $\uparrow$ & \textbf{Heading MAE} (rad) $\downarrow$ \\
        \midrule
        \rowcolor{gray!15} \multicolumn{5}{l}{\textit{Impact of Training Data Scaling (with Physics-Informed Hybrid Loss)}} \\
        Small (4.5k) & 2.08 & 5.43 & 20.60\% & 0.073 \\
        Medium (20k) & 1.51 & 3.92 & 27.14\% & 0.046 \\
        Large (40k) & 1.47 & 3.81 & 29.37\% & 0.044 \\
        \textbf{Standard (50k)} & \textbf{1.17} & \textbf{3.06} & \textbf{33.19\%} & \textbf{0.035} \\
        \midrule
        \rowcolor{gray!15} \multicolumn{5}{l}{\textit{Impact of Loss Components (50k Training dataset)}} \\
        CE & 1.47 & 3.82 & 29.00\% & 0.043 \\
        CE + REG & 1.21 & 3.17 & 32.08\% & 0.036 \\
        CE + REG + PIKC  & \textbf{1.17} & \textbf{3.06} & \textbf{33.19\%} & \textbf{0.035} \\
        \bottomrule
    \end{tabular}
    }
\end{table}

\subsection{Experiment 1: Fine-tuning StyleVLA on BEV Domain}
\label{subsec:experiment_1}
\subsubsection{Ablation Study}
\label{subsec:full_loss_ablation}

\textbf{Data Scaling Analysis}.
We trained the baseline Qwen2.5-VL-7B model on four subsets of our training dataset to determine the data volume required for robust generalization. The subsets were stratified by the number of distinct scenarios: small (4.5k samples), medium (20k samples), large (40k samples), and standard (50k samples). All models were trained with a fixed LoRA rank of 256.
\autoref{tab:ablation_combined} shows that increasing dataset size consistently improves performance. The standard dataset achieves the lowest \gls{ade} (\SI{1.17}{\meter}) and highest \gls{psr} (\SI{33.19}{\percent}), significantly outperforming the small subset (\SI{2.08}{\meter}, \SI{20.60}{\percent}), justifying our use of the standard dataset for final training.

\textbf{Loss function Analysis:}
To quantify the impact of our physics-informed hybrid loss framework as stated in \autoref{sec:finetuning}, we perform an ablation study on the Qwen2.5-VL-7B model trained on the standard dataset (50k), comparing three configurations: (1) \textit{CE} denotes $\mathcal{L}_\mathrm{ce}$; (2) \textit{CE + REG} corresponds to the hybrid loss $\mathcal{L}_\mathrm{hybrid}$; and (3) \textit{CE + REG + PIKC}  enhanced $\mathcal{L}_\mathrm{hybrid}$ by adding $\mathcal{L}_\mathrm{kin}$.
\autoref{tab:ablation_combined} highlights the progressive gains. Adding the \gls{mlp} regression head reduces the \gls{fde} by \SI{0.65}{\meter} ($\SI{3.82}{\meter}\to\SI{3.17}{\meter}$) and boosts the \gls{psr} by \SI{3.08}{\percent}. The kinematic consistency loss further refines control: the \gls{ade} improves ($\SI{1.21}{\meter}\to\SI{1.17}{\meter}$) and the \gls{psr} sees a gain (\SI{1.11}{\percent}), while the \gls{fde} improves ($\SI{3.17}{\meter}\to\SI{3.06}{\meter}$) and the Heading MAE is reduced ($\SI{0.036}{\radian}\to\SI{0.035}{\radian}$), confirming $\mathcal{L}_\mathrm{kin}$ as a vital physics-informed constraint.
\begin{table}[t]
    \centering
    \caption{Benchmarking across \glspl{vlm} on \gls{bev} domain (Zero-Shot)}
    \label{tab:bev_comparison}
    \resizebox{\columnwidth}{!}{%
    \renewcommand{\arraystretch}{1.1}
    \begin{tabular}{l ccccccc}
        \toprule
        \textbf{Model} & \textbf{Score} $\uparrow$ & \textbf{PSR} $\uparrow$ & \textbf{MR} $\downarrow$ & \textbf{ADE} $\downarrow$ & \textbf{FDE} $\downarrow$ & \textbf{KCE} $\downarrow$ & \textbf{Time} $\downarrow$ \\
        & (0-1) & (ADE$<1$m) & (FDE$>2$m) & (m) & (m) & (m) & (s) \\
        \midrule
        \rowcolor{gray!15} \multicolumn{8}{l}{\textit{Open Source Models}} \\
        Qwen3-VL-4B  & 0.00 & 0.00\% & 100.0\% & - & - & - & 2.00 \\
        Qwen2.5-VL-7B & 0.00 & 0.00\% & 100.0\% & - & - & - & 3.43 \\
        InternVL3-9B & 0.00 & 0.00\% & 100.0\% & 12.63 & 25.89 & 3.72 & 5.77 \\
        \midrule
        \rowcolor{gray!15} \multicolumn{8}{l}{\textit{Proprietary Models}} \\
        Gemini-2.5-Flash & 0.26 & 13.30\% & 73.40\% & 2.40 & 5.70 & 0.07 & 44.18 \\
        Gemini-2.5-Pro & 0.27 & 13.41\% & 71.57\% & 2.25 & 5.74 & 0.09 & 44.77 \\
        Gemini-3-Pro & 0.32 & 16.38\% & 66.21\% & 1.72 & 4.37 & 0.11 & 73.83 \\
        \midrule
        \rowcolor{gray!15} \multicolumn{8}{l}{\textit{Ours (Fine-Tuned)}} \\
        Qwen2.5-VL-7B & 0.46 & 33.19\% & 48.18\% & 1.17 & 3.06 & 0.12 & 3.7 \\
        \textbf{Qwen3-VL-4B} & \textbf{0.55} & \textbf{39.47\%} & \textbf{39.91\%} & \textbf{1.15} & \textbf{2.93} & \textbf{0.08} & \textbf{1.92} \\
        \bottomrule
    \end{tabular}
    }
    \par
    \vspace{2pt}
    \raggedright \footnotesize 
    Metrics are computed on generated trajectories. InternVL3-9B achieves a \SI{90.91}{\percent} generation rate; Qwen3-VL-4B and Qwen2.5-VL-7B generate none. “–” indicates metrics not computable due to zero success rate.
\end{table}
\subsubsection{Benchmarking across VLMs}
We fine-tuned two agents, Qwen2.5-VL-7B and Qwen3-VL-4B, on a training set of 50k samples (\SI{3}{\second}) and evaluated them on a held-out set of $2000$ samples. As summarized in \autoref{tab:bev_comparison}, we compared our framework against the high-performance open-source models and proprietary models in a Zero-Shot setting.

The results highlight three critical observations. First, \textbf{baseline open-source models fail completely} to generate valid driving-style trajectories (\SI{0}{\percent} success), showing that driving physics is not innate to standard pre-training. Second, \textbf{proprietary models} like Gemini-3-Pro, while performing best among baselines, still \textbf{struggle with precise trajectory generation} and require over \SI{70}{\second} per inference, making them unsuitable for online deployment.
Third, \textbf{fine-tuned models outperform even large-scale models}. Our fine-tuned version of Qwen3-VL-4B achieves a \SI{39.47}{\percent} success rate, while the best closed-source model achieves only \SI{16.38}{\percent}. Additionally, due to model size and quantization, the inference time remains online-capable (\SI{1.92}{\second}). This proves that domain-specific adaptation is essential for bridging the gap between reasoning and control. Furthermore, Qwen3-VL-4B demonstrates superior efficiency, achieving faster inference (\SI{1.92}{\second} vs. \SI{3.70}{\second}) and higher performance (Score $0.55$ vs. $0.46$) compared to the Qwen2.5-VL-7B model, aligning with technical reports~\cite{bai2025qwen3} that recent architectural advancements allow smaller models to match or exceed their predecessors. 
\begin{table}[t]
    \centering
    \caption{Benchmarking across \glspl{vlm} on \gls{fpv} domain (Zero-Shot)}
    \label{tab:carla_results}
    \resizebox{\columnwidth}{!}{%
    \renewcommand{\arraystretch}{1.1}
    \begin{tabular}{l ccccccc}
        \toprule
        \textbf{Model} & \textbf{Score} $\uparrow$ & \textbf{PSR} $\uparrow$ & \textbf{MR} $\downarrow$ & \textbf{ADE} $\downarrow$ & \textbf{FDE} $\downarrow$ & \textbf{KCE} $\downarrow$ & \textbf{Time} $\downarrow$ \\
        & (0-1) & (ADE$<1$m) & (FDE$>2$m) & (m) & (m) & (m) & (s) \\
        \midrule
        \rowcolor{gray!15} \multicolumn{8}{l}{\textit{Open Source Baselines}} \\
        Qwen3-VL-4B (Base) & 0.00 & - & - & - & - & - & 4.97 \\
        \midrule
        \rowcolor{gray!15} \multicolumn{8}{l}{\textit{Proprietary Models}} \\
        Gemini 2.5 Flash & 0.12 & 9.04\% & 87.21\% & 9.39 & 18.21 & 3.42 & 1.80 \\
        Gemini 2.5 Pro & 0.29 & 16.63\% & 70.74\% & 2.23 & 5.76 & 0.06 & 35.48 \\
        GPT-5 Nano & 0.29 & 16.67\% & 69.70\% & 2.36 & 6.06 & 0.11 & 49.05 \\
        Gemini 3 Pro & 0.35 & 17.65\% & 62.35\% & 1.54 & 3.94 & \textbf{0.06} & 91.39 \\
        \midrule
        \rowcolor{gray!15} \multicolumn{8}{l}{\textit{\gls{sota} Models}} \\
        SimLingo (1B) & 0.00 & 0.30\% & 99.40\% & 8.01 & 17.58 & / & 0.55 \\
        Orion (7B) & 0.05 & 2.10\% & 96.40\% & 11.13 & 21.35 & / & \textbf{0.36} \\
        OpenDriveVLA (0.5B) & 0.13 & 7.38\% & 87.25\% & 5.83 & 9.82 & / & 0.51 \\
        Alpamayo-R1 (10B) & 0.19 & 13.60\% & 73.10\% & 3.37 & 5.85 & / & 0.65 \\
        \midrule
        \rowcolor{gray!15} \textbf{Qwen3-VL-4B (StyleVLA)} & \textbf{0.51} & \textbf{38.60\%} & \textbf{36.90\%} & \textbf{1.17} & \textbf{3.13} & 0.11 & 2.13 \\
        \bottomrule
    \end{tabular}
    }
    \par
    \vspace{2pt}
   \raggedright \footnotesize
    Metrics are computed on successfully generated trajectories. All models achieve \SI{100}{\percent} generation except OpenDriveVLA (\SI{14.9}{\percent}), Gemini~2.5~Flash (\SI{90.9}{\percent}), and Qwen3-VL-4B (Base, \SI{0}{\percent}). - indicates metrics not computable due to zero success rate. 
/ indicates models lacking velocity or acceleration outputs; therefore, KCE cannot be computed.

\end{table}

\subsection{Experiment 2: Fine-tuning  StyleVLA on FPV Domain}
Based on the ablation study and \gls{bev} benchmarking results, we select Qwen3-VL-4B as the optimal backbone and employ the physics-informed hybrid loss to fine-tune our StyleVLA agent (based on 40k samples). The training hyperparameters are listed in \autoref{tab:loss_hparams}. We benchmark its end-to-end capabilities on the CARLA \gls{fpv} dataset ($1000$ samples) against high-impact pre-trained models. We also include \gls{sota} \gls{vla} models into this comparison, as they are specifically tailored for \gls{e2e} \gls{ad} and offer open-source checkpoints for reproducible evaluation. \autoref{fig:carla_image} provides qualitative comparisons across pretrained \glspl{vlm} and \gls{sota} baselines under five driving styles.

As shown in \autoref{tab:carla_results}, three trends emerge. First, among \textbf{proprietary models}, Gemini-3-Pro achieves the best zero-shot performance but \textbf{suffers from prohibitive latency} (\SI{91.39}{\second}), mirroring the \gls{bev} results in \autoref{tab:bev_comparison}. Second, baseline models and \gls{sota} \textbf{methods fail to generate high-quality driving-style trajectories}, likely due to insufficient fine-tuning on style-specific datasets. Furthermore, these \gls{sota} methods focus primarily on path generation and cannot output velocity or acceleration, making it impossible to evaluate their kinematic consistency. Third, \textbf{our fine-tuned agent significantly outperforms both} proprietary and \gls{sota} baselines in Score ($0.51$), PSR (\SI{38.60}{\percent}), and MR (\SI{36.90}{\percent}). This confirms the value of our StyleVLA dataset and demonstrates that open source lightweight \glspl{vlm}, after fine-tuning, can achieve competitive performance on domain-specific tasks, enabling effective \gls{e2e} driving-style trajectory generation. While our inference time (\SI{2.13}{\second}) is slightly higher than that of the \gls{bev} model (\SI{1.92}{\second}), this is expected as the \gls{fpv} agent must perform implicit perception to detect obstacles from raw images without explicit Traffic State lists (\autoref{sec:fpv_instruction}). Despite this additional complexity, the \gls{fpv} StyleVLA model trails the \gls{bev} StyleVLA model by only $\sim$\SI{0.9}{\percent} in PSR (\SI{38.60}{\percent} vs. \SI{39.47}{\percent}). 
\section{Conclusion and Future Work}
In this paper, we presented a comprehensive framework to generate the StyleVLA dataset, a large-scale instruction dataset (1.2k scenarios, 76k \gls{bev} and 42k \gls{fpv} samples) tailored for diverse driving styles (Default, Balanced, Comfort, Sporty, Safety). Based on the proposed dataset with \gls{bev} and \gls{fpv} visual contexts, we benchmarked the performance of pre-trained off-the-shelf \glspl{vlm} and \gls{sota} \gls{vla} models. We showed that even top proprietary models like Gemini-3-Pro fail to generate valid driving-style trajectories. 
We also fine-tuned \gls{vla} models with the Qwen3-VL-4B backbone using a physics-informed hybrid loss. As demonstrated by our ablation study, this approach significantly outperforms zero-shot \glspl{vlm} and \gls{sota} \gls{vla} models.
Our StyleVLA model achieves driving scores of $0.55$ (\SI{39}{\percent} success rate) in \gls{bev} and $0.51$ (\SI{38}{\percent} success rate) in \gls{fpv} with an average inference time of \SI{2}{\second}, surpassing the best baseline (Gemini-3-Pro), which scores only $0.32$ (\SI{16}{\percent} success rate) in \gls{bev} and $0.35$ (\SI{17}{\percent} success rate) in \gls{fpv} with high latency.
For future work, we aim to extend our framework with a novel action decoder to reduce inference time. We also plan to convert our StyleVLA simulation images into photorealistic images to improve the realism and fidelity of the data.

\bibliographystyle{IEEEtran}
\bibliography{literature.bib}


\end{document}